\documentclass[journal,twoside,web]{ieeecolor}
\usepackage{tmi}
\usepackage{cite}
\usepackage{amsmath,amssymb,amsfonts}
\usepackage[colorlinks,linkcolor=blue]{hyperref}
\usepackage{graphicx}
\usepackage{textcomp}
\usepackage{algorithm2e}
\RestyleAlgo{ruled}
\SetKwComment{Comment}{/* }{ */}
\usepackage{multirow}
\usepackage{colortbl}
\usepackage{booktabs}
\usepackage{tabulary}
\usepackage{xcolor}

\newcolumntype{C}[1]{>{\centering\arraybackslash}m{#1}}
\newcolumntype{?}[1]{!{\vrule width #1}}

\newcommand{\ylq}[1]{{\color{red}{[LQ: #1]}}}
\newcommand{\ylqq}[2]{{\color{gray}{#1}}{\color{red}{[#2]}}}
\def\ie{\emph{i.e.}}
\def\eg{\emph{e.g.}}
\def\etal{{\em et al.}}

\def\BibTeX{{\rm B\kern-.05em{\sc i\kern-.025em b}\kern-.08em
    T\kern-.1667em\lower.7ex\hbox{E}\kern-.125emX}}
\makeatletter
\def\endthebibliography{%
  \def\@noitemerr{\@latex@warning{Empty `thebibliography' environment}}%
  \endlist
}
\makeatother

\begin{document}
\title{Domain-incremental Cardiac Image Segmentation with Style-oriented Replay and Domain-sensitive Feature Whitening}
\author{Kang Li, Lequan Yu, and Pheng-Ann Heng
\thanks{
The work was supported in part by the Hong Kong Innovation and
Technology Fund (Project No. GHP/110/19SZ and No. ITS/170/20), and HKU Seed Fund for Basic Research (Project No. 202009185079 and 202111159073).
}
\thanks{K. Li and P.-A. Heng are with the Department of Computer Science
and Engineering, and the Institute of Medical Intelligence and XR at
The Chinese University of Hong Kong, HKSAR (e-mail: \{kli, pheng\}@cse.cuhk.edu.hk).}
\thanks{L. Yu is with the Department of Statistics and Actuarial Science,
The University of Hong Kong, HKSAR (e-mail: lqyu@hku.hk).}
}

\maketitle

\begin{abstract}
	%
	%
	%
	Contemporary methods have shown promising results on cardiac image segmentation, but merely in static learning, \ie, optimizing the network once for all, ignoring potential needs for model updating.
	%
	%
	In real-world scenarios, new data continues to be gathered from multiple institutions over time and new demands keep growing to pursue more satisfying performance.
	%
	%
	%
	The desired model should incrementally learn from each incoming dataset and progressively update with improved functionality as time goes by.
	%
	%
	As the datasets sequentially delivered from multiple sites are normally heterogenous with domain discrepancy, each updated model should not catastrophically forget previously learned domains while well generalizing to currently arrived domains or even unseen domains.
	%
	%
	%
	In medical scenarios, this is particularly challenging as accessing or storing past data is commonly not allowed due to data privacy.
	%
	%
	%
	%
	To this end, we propose a novel domain-incremental learning framework to recover past domain inputs first and then regularly replay them during model optimization. 
	%
	Particularly, we first present a style-oriented replay module to enable structure-realistic and memory-efficient reproduction of past data, and then incorporate the replayed past data to jointly optimize the model with current data to alleviate catastrophic forgetting.
	%
	%
	During optimization, we additionally perform domain-sensitive feature whitening to suppress model's dependency on features that are sensitive to domain changes (\eg, domain-distinctive style features) to assist domain-invariant feature exploration and gradually improve the generalization performance of the network.
	%
	We have extensively evaluated our approach with the M\&Ms Dataset in single-domain and compound-domain incremental learning settings.
	Our approach outperforms other comparison methods with less forgetting on past domains and better generalization on current domains and unseen domains.

\end{abstract}


\section{Introduction}
According to recent studies, cardiovascular diseases have become the leading cause of death. The mortality rate of cardiovascular diseases increases progressively year after year~\cite{chen2020deep}.
%
One fundamental task for computer-aided cardiovascular disease diagnosis is cardiac image segmentation, which delineates heart structures for many non-invasive volumetric quantifications, including stroke volumes, ejection fraction, myocardium thickness, and etc~\cite{chen2020deep,bala2020deep}.
With recent advances in deep learning, considerable deep learning-based approaches have achieved promising performance in this task~\cite{chen2020deep, bernard2018deep, zhuang2019evaluation}.
%
%
However, most of them adopt the static learning setup, where they assume model training and data delivery as a one-step process, but ignore the potential need for model upgrading~\cite{gonzalez2020wrong,li2017learning,yoon2021federated}.
%
In real-world clinical scenarios, new patients keep turning up day by day and induce new data samples to \emph{gather over time across institutions}.
This provides sufficient possibility to promote model functionality towards perfection.
To reach this goal, incremental learning (IL) is essential to investigate, which encourages the model to consecutively update upon the previously learnt one to progressively improve itself by exploiting each incoming dataset over time~\cite{bernard2018deep,perkonigg2021continual,gonzalez2020wrong}.
As a matter of fact, the datasets sequentially arrived from multiple institutions are normally heterogeneous with considerable domain shifts.
In this case, for each update in incremental learning lifespan, the network is prone to concentrate too much on adapting to the current domain, while disrupting the knowledge learnt from previous domains.
This would easily lead to huge performance degradation on past domains, especially considering that deep models are notoriously sensitive to the domain inconsistency between training data and test data.
%
This phenomenon is referred as Catastrophic Forgetting (CF) in recent literature~\cite{li2017learning}.
How to \emph{prevent catastrophic forgetting on previously learnt domains} while \emph{well accommodating to currently delivered domains or even unseen domains} is the pivotal challenge in domain-incremental learning.
Particularly, due to data privacy in medical areas~\cite{qu2021handling,yoon2021federated}, in each time step, the network only has the access to the current dataset and no previous dataset.
%
It is normally acceptable to pass the trained model among each other, but the data itself is strictly forbidden.
As a result, the model trained on past datasets is free to use, but the data samples are no longer accessible.
Furthermore, the network should not be aware of domain identity (\eg, the institution or vendor that each data sample comes from) in neither the training nor inference stage.
If domain identity is highly participated in training (e.g., developing customized strategies or parameters for each domain~\cite{liu2020ms, gonzalez2020wrong}), the model could be restricted to only be feasible to the domains that have been seen before.
It is also less practical to assume this information is available during inference, since the domain identity is possibly erased along with other patient privacy during data anonymization.

To conduct domain-incremental learning, the straightforward approach is to sequentially finetune the model with each incoming dataset.
However, with the absence of past domain inputs, directly finetuning the model by current heterogenous domain would easily disrupt the domain-specific embedding learnt for previous domains and cause severe forgetting on them~\cite{abraham2005memory,kemker2018measuring}.
%
%
%
%
Being aware of that, a handful of works attempted to record past domain information by storing a subset of data samples~\cite{perkonigg2021continual,rebuffi2017icarl,guo2020improved} or representative features~\cite{farajtabar2020orthogonal,hofmanninger2020dynamic}, and regularly replay them to help the model remember old knowledge.
%
%
Unfortunately, these approaches are infeasible in medical areas due to strict data privacy policy, which prohibits unregulated storage and transfer of data or features~\cite{qu2021handling,yoon2021federated}.
%
%
%
%
%
%
Alternatively, generative replay methods~\cite{shin2017continual}  constructed pseudo image-label pairs to simulate past domain inputs, where they synthesized pseudo-images from random noise by generative models and regarded their predictions of the previous model as pseudo-labels.
%
%
Guided by the pseudo image-label pairs, the model would less suffer from catastrophic forgetting.


%
%
%

However, several limitations impede the efficiency of the aforementioned approach.
First, as the pseudo image-label pairs generated by the existing generative replay strategy~\cite{shin2017continual} still have considerable gap to reach the precision of real image-label pairs, these would inevitably bring unreliable guidance to mislead the model.
%
%
%
Since they generated pseudo-images from random noise while the generative adversarial networks mainly concentrate on mimicking global distribution, the synthesized images may blend with artificial and unrealistic structures, harming the reliability of content embedding.
%
Less realistic pseudo-images would easily induce less reliable pseudo-labels and their combination would provide error-prone supervisions to misguide the model.
%
%
Second, the above strategy causes huge computational burden and memory consumption, as the replay of each past domain requires to train and store one specific generative network.
%
%
Last but not the least, this strategy hardly allows generalization to domains that could be encountered in the future.
Simply finetuning the model would let the feature embedding partially rely on the style representations of previously seen domains.
When it comes to the domain with unseen style, the model could be vulnerable and produce less satisfactory predictions.
To tackle these limitations, in this paper, we present a novel domain-incremental learning framework to enable progressive updates of cardiac image segmentation model with effective exploration of the sequentially arrived heterogenous data from multiple sites.
%
%
%
%
In each time step, the updated model does not catastrophically forget past domains and can well adapt to current domains and unseen domains.
%
%
To achieve it, we follow the previous generative replay strategy to reproduce previous domains first and then replay them during model optimization.
%
Differently, we recover past domain inputs in a structure-realistic and memory-efficient way via the proposed style-oriented replay module, and additionally present domain-sensitive feature whitening to facilitate the generalization on current domains and broaden its generalizable potential to unseen domains.
Specifically, our style-oriented replay module consists of (1) a base generative model whose parameters are trained by the first arrived domain, and (2) a style bank to record particular style adjustments upon the base generative model in mimicking the domains successively delivered after the first one.
%
%
%
Compared to maintaining a completely new generative model for each domain, conducting lightweight style-oriented modulation on top of the shared base generative model could maximally reuse the parameters learnt before.
%
This strategy not only alleviates the difficulties in generative replay, but also largely reduces unnecessary memory consumption.
Moreover, to guarantee realistic structure embedding of replayed images, we encourage the base generative model to embed the structures of segmentation labels into the replayed images via a pixel-wise paired image translation model (\eg, CGAN~\cite{isola2017image}), instead of generating them from random noise.
Then, we pair each replayed image with corresponding segmentation label as the recovered past inputs.
%
%
%
To alleviate catastrophic forgetting, we merge the replayed past data with current data to co-optimize the model.
Importantly, we further conduct domain-sensitive feature whitening to suppress the interference of the representations that are highly responsive to domain shifts (\eg, domain-specific style features), and let the model explore more on domain-invariant content features, guiding it to optimize towards a more generalizable direction.

We extensively validated our method on M\&Ms dataset~\cite{campello2021multi} in single-domain and compound-domain incremental learning settings.
Our approach outperforms other comparison methods with less forgetting on past domains and more improvements on current and unseen domains.
Our main contributions could be summarized as follows:
\begin{itemize}
	\item We present a novel domain-incremental learning framework to progressively update cardiac image segmentation model by investigating sequentially-arrived heterogeneous data.
	\item We propose the style-oriented replay module to enable structure-realistic and memory-efficient reproduction of past domains to mitigate catastrophic forgetting.
	\item We further propose to suppress the interference of domain-sensitive embedding by feature whitening to help generalize to newly-arrived and unseen domains.
	\item We have validated our framework on the M\&Ms dataset, where our method outperforms previous methods with less forgetting on past domains and better positive transfer on current and unseen domains.
	
\end{itemize}

\begin{figure*}[t]
	\centering
	\includegraphics[width=0.98\textwidth]{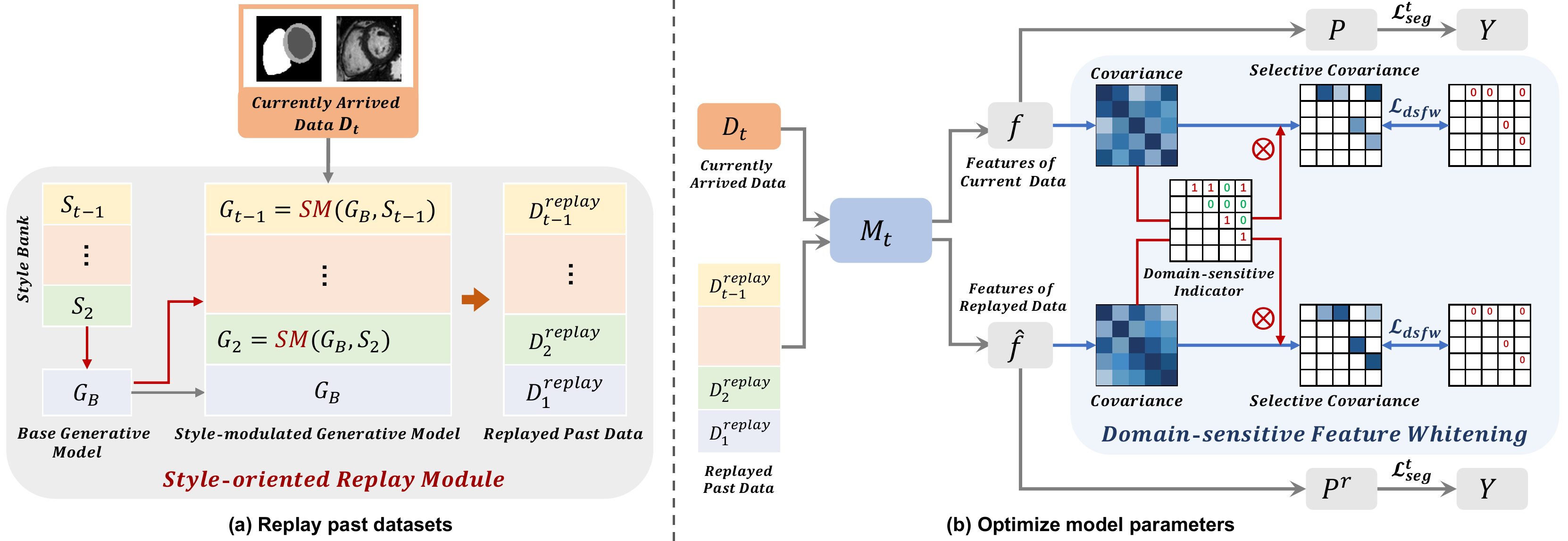}
	
	\caption{Overview of our framework. 
		With the arrival of the dataset $D_t$, our framework first conducts structure-realistic and memory-efficient recovery of past arrived datasets by our style-oriented replay module.
		Then, we invite the currently arrived dataset to co-optimize the model with replayed past datasets to mitigate catastrophic forgetting, while performing domain-sensitive feature whitening in the meantime to facilitate the generalization to current domains and unseen domains.
		(a): For memory-efficient replay, we greatly reuse the parameters of the base generative model $G_B$ previously trained on $D_1$ and conduct style modulation (SM) upon $G_B$ with trainable style parameters to stylize it oriented to each subsequent dataset after the first. Compared to storing one entire generative model for each dataset, saving only style modulation parameters in the style bank can greatly reduce unnecessary memory consumption.
		(b): When optimizing model parameters with the current data and replayed data, besides forcing the predictions of current data $P$ and that of replayed data $P^r$ to be consistent with the ground truth $Y$, we additionally conduct domain-sensitive feature whitening with the loss function $\mathcal{L}_{dsfw}$ to suppress the interference of the features that sensitively respond to domain changes (\eg, domain-distinctive style features) to broaden its generalization potential.}
	
	\label{fig:method}
\end{figure*}

\section{Related work}

%
%


%
\subsection{Cardiac Image Segmentation}
Cardiac image segmentation is a prerequisite task in computer-aided cardiovascular disease diagnosis, which supports multiple clinical quantitative measurements to assess the functional stage of heart~\cite{bala2020deep,chen2020deep}.
Until now, considerable efforts have been devoted to this task with promising performance by leveraging various powerful deep-learning architectures, such as 3D convolutional neural network~\cite{bala2020deep} or skip connection~\cite{ronneberger2015u} in single modality~\cite{zhuang2020cardiac,chen2020deep} or multiple modalities~\cite{zhuang2019evaluation}.
%
%
%
In general, these approaches train the network once for all in static learning, but barely consider underlying demands for updating.
In this case, incremental learning is undoubtedly essential, which enables consecutive training upon previously learned model to gradually perfect its functionality by exploiting each incoming dataset~\cite{bernard2018deep,perkonigg2021continual,gonzalez2020wrong}.
\subsection{Incremental Learning in Medical Image Analysis}
%
%
Several works have applied incremental learning in medical applications.
One stream of works~\cite{ozdemir2018learn,vu2021data,liu2021incremental} followed the problem settings in general computer vision~\cite{cermelli2020modeling}, which expect to build a class-incremental learning model over time, \ie, learn to segment new semantic structures while not forgetting previously learnt ones so that it would incrementally segment more and more structures.
%
%
%
%
In contrast, another stream~\cite{gonzalez2020wrong,ozgun2020importance,baweja2018towards,hofmanninger2020dynamic}, including ours, targeted on domain-incremental learning to continually improve the model towards a more robust and generalizable direction over time.
Specifically, it learned to segment certain semantic structures all along, and focused on how to incrementally train the model to accommodate more domains or even unseen domains while not forgetting earlier acquired ones.
%
%

Due to particular constraints for medical applications, several approaches are normally infeasible to apply in real-world scenarios.
%
%
For example, the intuitive method, \ie, incrementally joint training the model by all arrived datasets, is impractical as previously delivered data are not accessible anymore due to data privacy concerns~\cite{qu2021handling,yoon2021federated,li2021fedbn}.
This privacy regulation also limits the feasibility of several replay-based incremental learning methods, especially for those intending to store a subset of data~\cite{perkonigg2021continual} or features~\cite{hofmanninger2020dynamic} of previously seen domains into a separate memory unit to avoid severe forgetting on them.
%
%
Additionally, the existing works that took domain identity as extra training or inference guidance to design specific treatment for each domain~\cite{karani2018lifelong} is also infeasible, due to data anonymity procedures.
%
%
%

To prevent catastrophic forgetting on past domains, several approaches employed a regularization term, such as elastic weight consolidation~\cite{baweja2018towards} and memory aware synapses~\cite{ozgun2020importance}, to constrain large updates of the parameters that are important to segment old domains.
However, as pointed out by Saha \etal~\cite{saha2020gradient}, the performance of these methods suffer from learning longer task sequence.
%
Gonzale \etal~\cite{gonzalez2020wrong} sought another way which trained and stored an individual model for each domain to completely isolate the parameters.
During inference, they adopt a VAE-based domain classifier to determine which model to retrieve and test.
Undoubtedly, it causes excessive memory consumption which increases linearly as time goes on.
Several works~\cite{mallya2018packnet,mallya2018piggyback} got inspiration from network quantization and pruning, where they
proposed to prevent catastrophic forgetting by isolating the parameters of old domains or tasks and leveraging learnable piggyback masks~\cite{mallya2018piggyback} or iterative parameter pruning~\cite{mallya2018packnet} during inference.
Despite many merits of them, \eg, agnostic to task order and improved robustness and performance, these works could be vulnerable to long sequences of tasks, considering the knowledge capacity of deep learning models is bounded~\cite{hu2021model}.
Some recent works proposed to mitigate catastrophic forgetting by regularly replaying the old domain knowledge, such as the outputs of previous model~\cite{li2017learning}, or imitated past domain inputs~\cite{shin2017continual}.
Our framework also follows this strategy, while we intend to reproduce previous inputs in a structural-authentic and memory-efficient way with the proposed style-oriented replay module.
Importantly, we further facilitate the generalization on new domains and unseen domains by a new domain-sensitive feature whitening.


\section{Methodology}
%
%
%
The datasets sequentially arrived over time across multiple institutions are normally heterogenous with considerable domain shifts.
%
%
In time step $t$, the previous model $M_{t-1}$ has been trained by a stream of heterogenous datasets delivered earlier $\mathcal{D}^{p}_{t}=\{D_{1}, D_{2}, ..., D_{t-1}\}$ in sequence.
%
%
%
%
%
With the arrival of the dataset $D_{t}$, our target is to incrementally optimize the model $M_{t}$ based on $M_{t-1}$ such that the updated model $M_t$ would (1) not catastrophically forget previously learnt domains $\mathcal{D}^{p}_{t}$, and (2) well adapt to the current domain $D_{t}$ or even unseen domains.
%
It is worth noting that only the newly incoming dataset $D_{t}$ is accessible, while the past domain inputs $\mathcal{D}^{p}_{t}$ are no longer available due to data privacy concerns.
Moreover, no domain identity (\eg, dataset or institution identity) is provided in either the training or inference stage.

Fig.~\ref{fig:method} illustrates our proposed domain-incremental learning framework.
%
%
Our framework is based on generative replay  strategy.
We first recover past datasets in a memory-efficient and structure-realistic way by a novel style-oriented replay module, and then regularly remind the model with the replayed past datasets to alleviate catastrophic forgetting.
%
Upon that, we propose to additionally conduct domain-sensitive feature whitening during optimization, aiming to improve the generalization on current data and unseen data by suppressing model dependency on the representations that sensitively respond to domain changes (\eg, domain-specific style features) and implicitly promoting the exploration on domain-invariant features.
%
%
The details are elaborated in the following subsections.
%

%
%

\subsection{The Style-oriented Replay Module}
\begin{figure}[t]
	\centering
	\includegraphics[width=0.48\textwidth]{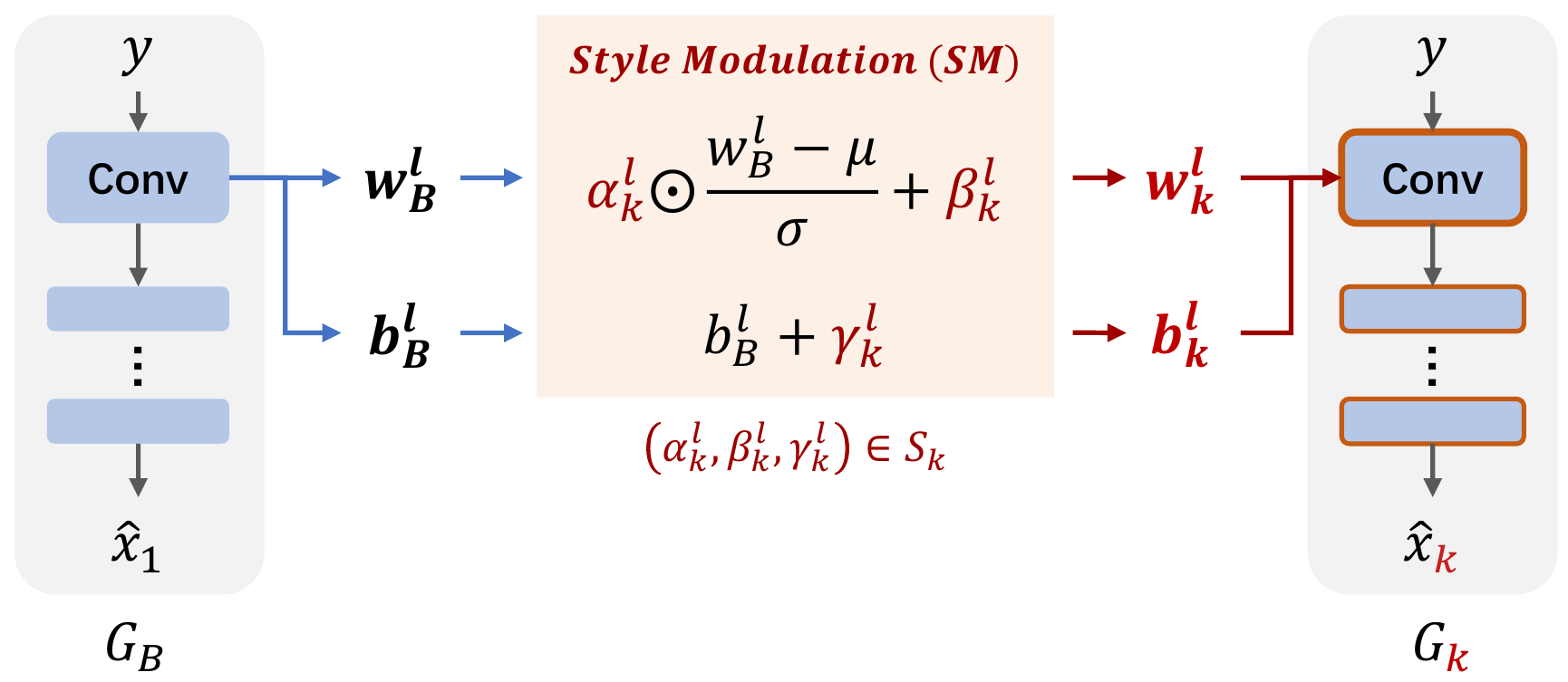}
	\caption{An example of style modulation from $G_B$ to $G_k$ at the $l$-th layer.}
	\label{fig:method-style}
\end{figure}

Given the first arrived dataset $D_1$, the model $M_1$ is trained under full supervision of the annotations in $D_1$ by the objective function $\mathcal{L}_{sup}^{1}$, which is formed as
%
%
%
%
\begin{equation}
	\mathcal{L}_{sup}^{1} = \mathcal{L}_{ce}(M_1(X_1), Y_1),
	\label{eq:sup-1}
\end{equation}
where $\mathcal{L}_{ce}$ presents the cross-entropy loss, and $X_1$, $Y_1$ denote the sets of images and annotations of $D_1$ respectively.
For any subsequent time step $t$, \ie, $t\in[2, T]$, the issue of catastrophic forgetting comes up.
To address it, our framework presents the style-oriented replay module to simulate all past datasets $\mathcal{D}^{p}_{t}$ by generative models as $\mathcal{D}^{r}_{t}= \{D_{1}^{replay}, D_{2}^{replay}, ..., D_{t-1}^{replay}\}$.
%
%
Then, we invite the replayed old data $\mathcal{D}^{r}_{t}$ to co-optimize the model $M_{t}$ along with currently arrived data $D_{t}$ to help retain previously learnt knowledge.
%
%
%
%
%

Our style-oriented replay module consists of one base generative model $G_B$ trained on the first dataset $D_{1}$, and a style bank $\mathcal{S}$ to record the style adjustment parameters on top of $G_B$ when simulating the distribution of each dataset delivered after the first one.

\subsubsection{The base generative model}
To encourage realistic structure embedding in replayed images, we implement the base generative model $G_B$ following paired image translation networks (\eg, CGAN~\cite{isola2017image}), such that the structures beneath the replayed images would come from the segmentation labels under pixel-to-pixel mapping.
%
We equip $G_B$ with a discriminator network $F_B$ to form an adversarial game against each other.
Given any label-image pair $(y, x)$ of $D_{1}$, the generator takes segmentation label $y$ as input to recover the corresponding image $x$ as realistic as possible, while the discriminator $F_B$ aims to identify the real image $x$ from the generated one $\hat{x}=G_B(y)$.
The objective function $\mathcal{L}_{adv}(G_B, F_B)$ is computed as
\begin{equation}
	\begin{aligned}
		\mathcal{L}_{adv}(G_B, F_B) =~& \mathbb{E}_{(y, x)\in D_{1}}[\log F_B(y,x)]+ \\
		& \mathbb{E}_{y\in D_{1}}[\log (1-F_B(y, G_B(y))].
	\end{aligned}
	\label{eq:base-G}
\end{equation}

\subsubsection{The style bank}
For each dataset $D_{i}~(i\in[2, t])$ arrived after the first one, we propose to efficiently reuse the parameters of the base generative model $G_B$ trained earlier, and tackle their style variations from $D_{1}$ by particular style modulation operations.
Recent work in style transfer literature~\cite{zhao2020leveraging} demonstrated that one can manipulate a generative model trained on one dataset (\eg, $D_1$) to mimic the distribution of another dataset (\eg, $D_{i}$) by conducting element-wise affine transformation on filters.
Inspired by that, we adjust the filters of the base generative model $G_B$ with trainable style modulation parameters $S_{i}$ to stylize it oriented to $D_i$.
Take the example of the $l$-th Conv layer in $G_B$ with $H_{out}^{l} = \mathbf{W}_{B}^{l}H_{in}^{l} + \mathbf{b}_{B}^{l}$, we denote the filters as $\mathbf{W}_{B}^{l}\in \mathbb{R}^{C_{out}\times C_{in} \times K_{1} \times K_{2}}$, bias as $\mathbf{b}_{B}^{l} \in \mathbb{R}^{C_{out}}$,  and the input and output features as $H_{in}^{l}$ and $H_{out}^{l}$ respectively.
As shown in Fig.~\ref{fig:method-style}, we freeze $(\mathbf{W}_{B}^{l}, \mathbf{b}_{B}^{l})$ and adjust them with trainable parameters $S_i^{l} = (\alpha_{i}^{l}, \beta_{i}^{l}, \gamma_{i}^{l})$ to modulate its style to $D_i$ as:
%
%
%
\begin{equation}
	\begin{aligned}
		\mathbf{W}_{i}^{l} &= \alpha_{i}^{l} \odot \frac{\mathbf{W}_{B}^{l}-\mu}{\sigma}+\beta_{i}^{l},\\
		\mathbf{b}_{i}^{l} &= \mathbf{b}_{B}^{l}+\gamma_{i}^{l},
	\end{aligned}
	\label{eq:style-modulate}
\end{equation}
where $\alpha_{i}^{l}, \beta_{i}^{l} \in \mathbb{R}^{C_{out}\times C_{in}}$,  $\gamma_{i}^{l} \in \mathbb{R}^{C_{out}}$ and $\odot$ denotes the Hadamard product.
%
$\mu, \sigma \in \mathbb{R}^{C_{out}\times C_{in}}$ denote the mean and standard deviation of $\mathbf{W}_{B}^{l}$ respectively.
%
%
The above style modulation is performed for each Conv layer, then the generator and discriminator for $D_{i}$ are formulated as $G_{i} = \{(\mathbf{W}_{i}^{l}, \mathbf{b}_{i}^{l}), l\in[1, N_{G}]\}$ and $F_{i} = \{(\mathbf{W}_{i}^{l}, \mathbf{b}_{i}^{l}), l\in[1, N_{F}]\}$ respectively, where $N_{G}$ and $N_{F}$ denote the total number of Conv layers in $G_B$ and $F_B$ accordingly.
%
%
Then, we optimize $G_{i}$ and $F_{i}$ with trainable parameters $S_{i}$ to mimic the distribution of $D_i$ by adversarial training as follows:
\begin{equation}
	\begin{aligned}
		\mathcal{L}_{adv}(G_{i}, F_{i}) =~& \mathbb{E}_{(y, x)\in D_{i}}[\log F_{i}(y,x)]+ \\
		& \mathbb{E}_{y\in D_{i}}[\log (1-F_{i}(y, G_{i}(y))].
	\end{aligned}
	\label{eq:style-G}
\end{equation}

For future replay of all previously seen datasets, we save the base generative model $G_{B}$ and store a stream of style modulation parameters $\{S_{2}, ..., S_{t-1}\}$ in the style bank $\mathcal{S}$.
Compared to train and save one entire generative model for each dataset, only adjust and store style modulation parameters greatly lighten the computational burden and memory cost.
%
%
%
%
%
%
%

With the arrival of each dataset $D_{t}$, we feed its segmentation label set $Y_t$ into the base generative model $G_B$ to replay the first dataset $D_1$ as $D_{1}^{replay}$, and modulate $G_B$ with each style parameters $S_i$ as the generative model $G_i$ style-oriented to $D_i$, $i\in [2, t-1]$, by Eq.~\ref{eq:style-modulate} and input $Y_t$ to replay $D_i$ as $D_{i}^{replay}$ following:
\begin{equation}
	\begin{aligned}
		D_{1}^{replay}&=\{(\hat{x}_{1}, y) \mid \hat{x}_{1}=G_B(y), y\in Y_{t}\},\\
		D_{i}^{replay}&=\{(\hat{x}_{i}, y) \mid \hat{x}_{i}=G_{i}(y), y\in Y_{t}\}.\\
	\end{aligned}
	\label{eq:replay}
\end{equation}

Sequentially, all previously seen datasets would be replayed as $\mathcal{D}^{r}_{t}=\{D_{1}^{replay}, D_{2}^{replay}, ..., D_{t-1}^{replay}\}$.
%
%
To remind the model with previously acquired data, we take $\mathcal{D}^{r}_{t}$ to jointly optimize $M_t$ together with $D_t$.
We initialize the current model $M_{t}$ with the previous one $M_{t-1}$ and update it as follows:
\begin{equation}
	\mathcal{L}_{sup}^{t} = \mathcal{L}_{ce}(M_t(X_t), Y_t) +  \frac{\lambda_{r}}{t-1} \sum_{i=1}^{t-1} \mathcal{L}_{ce}(M_t(\hat{X}_i), Y_t),
	\label{eq:seg-replay}
\end{equation}
where $\lambda_{r}$ is a hyper-parameter to adjust the importance of two terms, and $X_t$ and $\hat{X}_i$ denote the images of current dataset $D_t$ and the $i$-th replayed dataset $D_{i}^{replay}$, respectively.

\subsection{Domain-sensitive Feature Whitening}
\begin{figure}[t]
	\centering
	\includegraphics[width=0.4\textwidth]{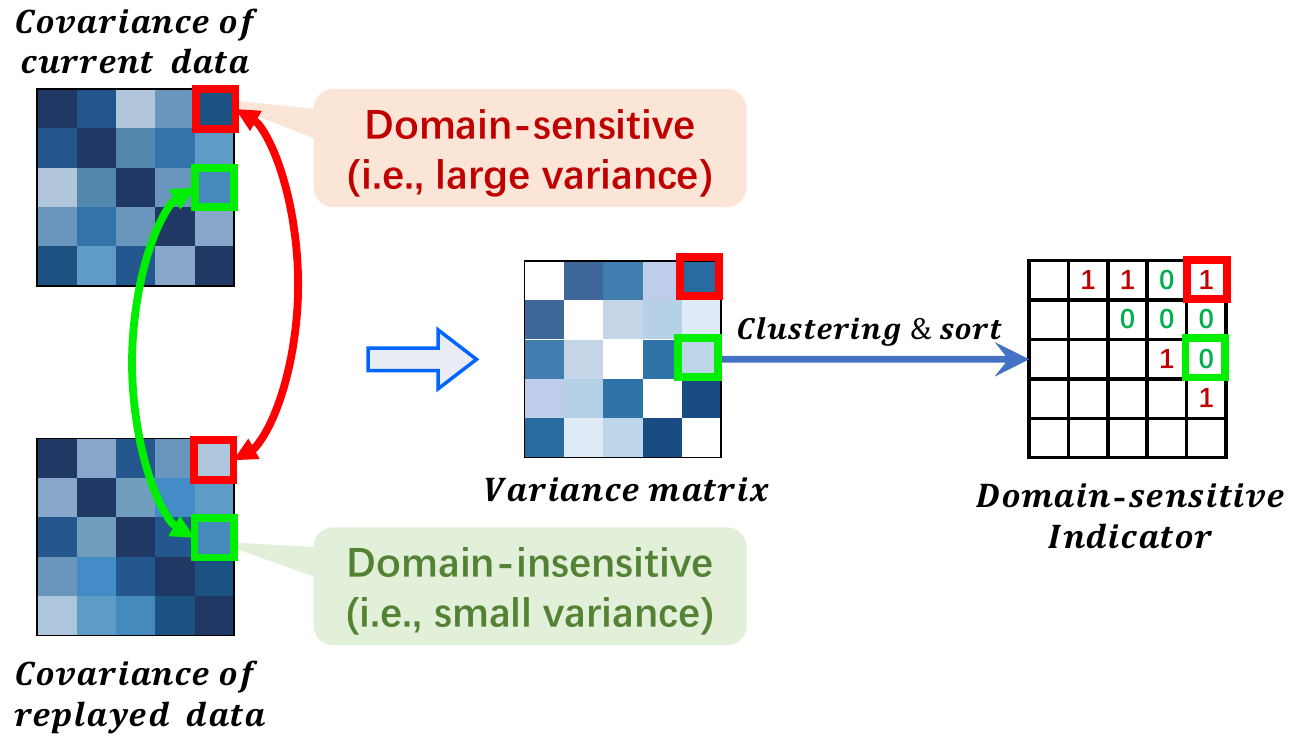}
	\caption{Illustration of identifying domain-sensitive parts. We explore the differences of high-order statistics (\ie, the variances of feature covariance metrics) between replayed and current features, and pick those with large variances (marked in red) as the domain-sensitive ones.}
	\label{fig:method-sensitive}
\end{figure}

To avoid hampering the model generalization potential when jointly training with replayed datasets $\mathcal{D}_t^r$ and current datasets $D_t$, we further suppress the model's dependency on the representations that are sensitive to domain changes, \eg, the domain-specific style features which are commonly considered as the main cause of domain shifts~\cite{qu2021handling}.
Implicitly, it encourages the model to explore deeper on the domain-invariant content features, and guides it towards a more generalizable direction.

%
%
%

Specifically, to highlight the feature variations brought by styles, we first regroup the elements of replayed datasets $\mathcal{D}^{r}_{t}$ and the current dataset $D_{t}$ as a new set $\mathcal{D}_{t} = \left\{(y^j, x^j, \hat{x}^j):(y^j, x^j) \in D_t,\hat{x}^j = \{\hat{x}_1^j, ..., \hat{x}_{t-1}^j\}, j \in [1, N_t]\right\}$, where $\forall i, i \in [1, t-1], \hat{x}_i^j = G_i(y^j)$.
In this way, each image in the set $[x^j, \hat{x}^j]$ would either correspond to or generate from the same segmentation label $y^j$, where they share the same content embedding, but distinguish to each other with style variations.
Then, we randomly select one replayed image $x_{rpl}^j$ from $\hat{x}^j$, and pair it with the corresponding input image $x^j$.
We input $x^j$ and $x_{rpl}^j$ into the same model $M_t$, and obtain the features $f^j$ and $\hat{f}^j$ at the same layer respectively, where $f^j, \hat{f}^j \in \mathbb{R}^{C\times HW}$.
%
Then, we conduct feature standardization on $f^j$ and $\hat{f}^j$ to scale each feature vector as the unit value by instance normalization~\cite{ulyanov2016instance} as
%
%
\begin{equation}
	\begin{aligned}
		f_s&=\left(\operatorname{diag}\left(\Sigma_{f}\right)\right)^{-\frac{1}{2}} * \left(f-\mu_{f}\right),\\
		\mu_{f}&=\frac{f}{H W},~
		\Sigma_{f}=\frac{1}{H W}\left(f-\mu_{f}\right)\left(f-\mu_{f} \right)^{\top},
	\end{aligned}
	\label{eq:fea-standard}
\end{equation}
where $*$ denotes the element-wise multiplication. Here, we denote the standardized features as $f_s^j$ and $\hat{f}_s^j$ respectively.
%
%
For each standardized feature, we calculate the channel-wise covariance matrix, where each element implies the dependency (\ie, correlation) between two channels as follows:
%
\begin{equation}
	\begin{aligned}
		\Sigma_s^j=\frac{1}{H W} (f_s^j) (f_s^j)^{\top},\\
		\hat{\Sigma}_s^j=\frac{1}{H W} (\hat{f}_s^j) (\hat{f}_s^j)^{\top},\\
	\end{aligned}
	\label{eq:fea-covar}
\end{equation}
where $\Sigma_s^j, \hat{\Sigma}_s^j \in \mathbb{R}^{C\times C}$.
%
%
%
%
%
As shown in Fig.~\ref{fig:method-sensitive}, since these two covariance matrices come from the features that share a similar content embedding but with style discrepancy, their element-wise variance changes would reflect the sensitivity triggered by domain-distinctive style changes.
%
%
%
Then, those with large variance could be considered as the domain-sensitive ones and vice versa.
%
To obtain an informative variance matrix, we randomly sample the image pair $(x^j, \hat{x}^j_i)$ for $N$ times and calculate the average variance matrix $V\in\mathbb{R}^{C\times C}$ as
\begin{equation}
	\begin{aligned}
		V&=\frac{1}{N} \sum_{j=1}^{N} \sigma_{j}^{2},\\
		\sigma_{j}^{2}&=\frac{1}{2}\left(\left(\Sigma_s^j-\mu_{\Sigma_{j}}\right)^{2}+\left(\hat{\Sigma}_s^j-\mu_{\Sigma_{j}}\right)^{2}\right),\\
		\mu_{\Sigma_{j}}&=\frac{1}{2}\left(\Sigma_s^j+\hat{\Sigma}_s^j\right),
	\end{aligned}
	\label{eq:fea-var}
\end{equation}
where $N$ is a hyper-parameter to indicate the total number of sampling.
Afterwards, we employ a simple but effective clustering technique, \ie, $k$-means, to separate the elements of $V$ into $k$ clusters $\{c_1, c_2, .., c_k\}$ and sort them in the ascending order.
Then, we include the items in the cluster of the largest variance $c_k$ into the high-variance group $C_{high}=\{c_k\}$, which implies the domain-sensitive representations, \eg, the domain-specific style features.
Meanwhile, we merge all left clusters into the low-variance group $C_{low}=\{c_1, ..., c_{k-1}\}$, which represents domain-insensitive representations, \eg, the domain-invariant content ones.
%
%
We use a matrix $I\in \mathbb{R}^{C\times C}$ to indicate the locations of domain-sensitive covariances as
\begin{equation}
	I_{i, j}=\left\{
	\begin{array}{ll}
		1, & \text { if } V_{i, j} \in C_{high} \\
		0, & \text { otherwise }
	\end{array}\right.
	\label{eq:fea-I}
\end{equation}
Then, we suppress those in the high-variance group by feature whitening~\cite{cho2019image} by $\mathcal{L}_{dsfw}$ loss, which minimizes the selected covariance to zero to decompose their correlation as follows:
%
%
\begin{equation}
	\mathcal{L}_{dsfw}=\mathbb{E}\left[\left\|\Sigma_{s} * I\right\|_{1}\right] + \mathbb{E}[\|\hat{\Sigma}_{s} * I\|_{1}].
	\label{sfw}
\end{equation}
%
It is worth to mention that the domain-sensitive feature whitening could be applied for any layer in the segmentation backbone.
It is also feasible to perform it with multiple-layer features.
%
In our case, we use the features after each max-pooling layer in the segmentation backbone, and calculate the average feature whitening loss.
%
The overall objective function to update the segmentation model $M_{t}$ is formulated as
\begin{equation}
	\mathcal{L}_{seg}^{t} =\mathcal{L}_{sup}^{t} + \lambda_{d} \mathcal{L}_{dsfw},
	\label{eq:overall}
\end{equation}
where $\lambda_{d}$ is a hyper-parameter to balance two terms.
Consequently, while alleviating catastrophic forgetting by using the first term, our framework could also promote the generalization to other domains with the help of the second term.


\subsection{Training Scheme}
With the arrival of the first batch of data ($t=1$), we first train a segmentation model $M_1$.
For future replay of $D_1$, we train a generative network to mimic its distribution and store it as the base generative model $G_B$ in the style-oriented replay module.
%
%
%
In each subsequent time step ($t>1$), we first reproduce all previously seen datasets $\mathcal{D}^r_{t} = \{D_1^{replay}, ..., D_{t-1}^{replay}\}$ via the style-oriented replay module, then employ the replayed data $\mathcal{D}^r_{t}$ to co-optimize current model $M_t$ with current dataset $D_{t}$ to avoid catastrophic forgetting on past datasets.
%
%
During optimization, we additionally perform domain-sensitive feature whitening for better generalization on the datasets that might be encountered in the future.
After updating the parameters of $M_t$, we use trainable parameters $S_t$ to conduct style modulation upon the base generative model $G_B$ to train a model style-oriented to the current data batch, and save the style modulation parameters $S_t$ in the style bank $\mathcal{S}$ for future replay.
The detailed training scheme is presented in Algorithm~\ref{Al:1}.
During the inference, only the current segmentation model $M_t$ is needed, and no domain identity (\ie, site or vendor identity) is required. 
%

\begin{algorithm}[hbt!]
	\caption{Training Procedure} 
	\label{Al:1}
	{\bf Output:} Segmentation model $\mathcal{M}$ at last time step $T$.\\
	\While{incrementally learning from $t=1$ to $T$}{
		\eIf{$t == 1$}{
			Train $M_1$ with $D_1 $ according to Eq.~\ref{eq:sup-1}.\\
			Train the base generative model $G_{B}$ to mimic the distribution of $D_1$ as Eq.~\ref{eq:base-G}. \\
			Save $G_{B}$ for future replay of $D_1$. \\
		}{
			Initialize $M_{t}$ by $M_{t-1}$. \\
			Replay previous datasets as $\mathcal{D}^r_{t}$ by the style-oriented replay module using Eq.~\ref{eq:replay}.\\
			Re-organize $D_t$ and $\mathcal{D}_{t}^r$ into $\mathcal{D}_t$.\\
			\For{$i=1$ to $N$}{
				Sample $(x, \hat{x})$ from $\mathcal{D}_t$\\
				Calculate the matrix $V$ by Eq.~\ref{eq:fea-var}.
			}
			Compute the indicator $I$ matrix by Eq.~\ref{eq:fea-I}.\\
			Optimize $M_t$ by replayed past data  $\mathcal{D}_t^r$ with domain-sensitive feature whitening by Eq.~\ref{eq:overall}.\\
			Conduct style modulation on $G_B$ with $S_t$ to mimic $D_{t}$ by Eq.~\ref{eq:style-modulate} and Eq.~\ref{eq:style-G}.\\
			Store $S_t$ into the style bank $\mathcal{S}$ for future replay of $D_t$ by Eq.~\ref{eq:replay}.\\
			
		}
		Pass $M_{t}$ to initialize the model in the $t+1$ step.\\
	}
	{\bf Return} $\mathcal{M} \leftarrow M_T$.
	
\end{algorithm}

\section{Experiment}
\begin{table*}[ht]
	\centering
	\caption{
		The Comparison Results to Incrementally Learn the Model by Single Domain Data. We First Reported the Performance of the Past, Current and Unseen Domains in the Final Time Step Respectively. Then We Presented the Performance During the Entire Incremental Learning Lifespan in the Last Five Columns. 
		For the results Evaluated by DSC, the Higher the Better. As For the Results Evaluated by HD, the Lower the Better. The Best Results Have Been Highlighted with \textbf{bold}}.
	
	\begin{tabular}{l|c|c|c|c|c|c|c|c|c} 
		\toprule[1pt]
		\multirow{2}{*}{Methods}                                               & \multicolumn{4}{c|}{DSC [\%]}                                                                                                 & \multirow{2}{*}{IL-DSC} & \multirow{2}{*}{BWT} & \multirow{2}{*}{TL} & \multirow{2}{*}{FTU} & \multirow{2}{*}{FTI}  \\ 
		\cline{2-5}
		& \multicolumn{1}{l|}{Vendor A} & \multicolumn{1}{l|}{Vendor B} & \multicolumn{1}{l|}{Vendor C} & \multicolumn{1}{l|}{Vendor D} &                         &                      &                     &                      &                       \\ 
		\midrule[1pt]
		Individual Training                                                    & 88.91                       & 89.82                       & 86.48                       & 88.13                      & 86.34\%                 & 96.03\%              & 88.40\%             & 83.87\%              & N/A                   \\ 
		\hline
		\begin{tabular}[c]{@{}l@{}}Joint Training\\ (Upper bound)\end{tabular} & 89.17                       & 90.22                       & 86.77                       & 86.24                       & 88.85\%                 & 99.82\%              & 88.63\%             & 83.93\%              & 0.38\%                \\ 
		\midrule
		Sequentially
		Finetune                                                & 84.11                       & 88.62                       & 86.71                       & 84.77                       & 87.56\%                 & 97.02\%              & 88.37\%             & 83.44\%              & 0.43\%                \\ 
		Orc-MML~\cite{gonzalez2020wrong}                                                                & 86.13                      & 87.39                       & 85.54                       & 85.05                      & 87.29\%                 & 98.06\%              & 88.12\%             & 83.31\%              & -1.14\%               \\ 
		MAS-LR~\cite{ozgun2020importance}                                                                & 84.89                       & 88.36                      & 86.57                       & 84.62                       & 87.43\%                 & 96.72\%              & 88.34\%             & 83.27\%              & -0.26\%               \\ 
		LwF~\cite{li2017learning}                                                                     & 85.42                       & 88.78                       & 86.22                       & 85.14                       & 87.64\%                 & 97.35\%              & 88.63\%             & 83.73\%              & 0.63\%                \\ 
		EWC~\cite{baweja2018towards}                                                                     & 84.74                      & 88.86                       & 86.79                       & 85.33                       & 87.76\%                 & 97.05\%              & 88.71\%             & 83.66\%              & 0.54\%                \\ 
		GR~\cite{shin2017continual}                                                                    & 85.11                       & 89.04                       & 86.44                       & 85.24                       & 87.84\%                 & 98.23\%              & 88.58\%             & 83.87\%              & -0.36\%               \\ 
		\midrule
		Ours w/o $L_{dsfw}$                                                         & 87.44                       & 89.57                       & 86.82                       & 85.36                       & 88.34\%                 & 99.07\%              & 88.81\%             & 83.52\%              & 0.37\%                \\ 
		Ours                                                                   & \textbf{87.95}                      & \textbf{89.73}                       & \textbf{87.51}                       & \textbf{86.17}                      & \textbf{88.72\%}                 & \textbf{99.36\%}              & \textbf{89.04\%}             & \textbf{84.45\%}              & \textbf{1.22\%}                \\
		\bottomrule[1pt]
	\end{tabular}
	\\[4pt]
	\begin{tabular}{l|c|c|c|c|c|c|c|c|c} 
		\toprule[1pt]
		\multirow{2}{*}{Methods}                                               & \multicolumn{4}{c|}{HD [Pixel]}           & \multicolumn{1}{c|}{\multirow{2}{*}{IL-HD}} & \multicolumn{1}{c|}{\multirow{2}{*}{$\text{BWT}^{+}$}} & \multicolumn{1}{c|}{\multirow{2}{*}{TL}} & \multicolumn{1}{c|}{\multirow{2}{*}{FTU}} & \multicolumn{1}{c}{\multirow{2}{*}{FTI}}  \\ 
		\cline{2-5}
		& Vendor A & Vendor B & Vendor C & Vendor D & \multicolumn{1}{l|}{}                       & \multicolumn{1}{l|}{}                      & \multicolumn{1}{l|}{}                    & \multicolumn{1}{l|}{}                     & \multicolumn{1}{l}{}                      \\ 
		\toprule[1pt]
		Individual Training                                                    & 14.89    & 10.05    & 11.99    & 15.50    & 11.25                                       & 3.93                                       & 9.77                                     & 13.64                                     & \multicolumn{1}{l}{N/A}                   \\ 
		\hline
		\begin{tabular}[c]{@{}l@{}}Joint Training\\ (Upper bound)\end{tabular} & 8.93     & 7.66     & 11.42    & 12.01    & 9.05                                        & 0.00                                       & 9.60                                     & 13.53                                     & -0.25                                      \\ 
		\midrule
		Sequentially Finetune                                                  & 12.99    & 8.92     & 11.16    & 14.10    & 10.49                                       & 2.65                                       & 9.51                                     & 13.57                                     & -0.28                                      \\ 
		
		Orc-MML~\cite{gonzalez2020wrong}                                                                & 11.39    & 8.74     & 13.14    & 13.76    & 10.37                                       & 1.42                                       & 10.33                                    & 13.74                                     & 0.85                                       \\ 
		
		MAS-LR~\cite{ozgun2020importance}                                                                 & 13.42    & 9.80     & 11.63    & 15.40    & 10.72                                       & 1.98                                       & 10.15                                    & 14.05                                     & 0.57                                       \\ 
		
		LwF~\cite{li2017learning}                                                                    & 14.05    & 9.64     & 11.25    & 15.54    & 10.85                                       & 3.37                                       & 9.53                                     & 13.62                                     & -0.36                                      \\ 
		
		EWC~\cite{baweja2018towards}                                                                    & 11.78    & 8.66     & 11.21    & 13.25    & 10.25                                       & 2.01                                       & 9.60                                     & 13.76                                     & -0.25                                      \\ 
		
		GR~\cite{shin2017continual}                                                                     & 12.07    & \textbf{8.32}     & 11.13    & 13.01    & 10.01                                       & 1.57                                       & 9.61                                     & 13.82                                     & 0.01                                       \\ 
		\midrule
		Ours                                                                   & \textbf{10.50}    & 8.63     & \textbf{10.56}    & \textbf{12.39}    & \textbf{9.62}                                        & \textbf{0.91}                                       & \textbf{9.29}                                     & \textbf{13.39}                                     & \textbf{-0.54}                                      \\
		\bottomrule[1pt]
	\end{tabular}
	\label{table:single}
\end{table*}
\subsection{Datasets and Implementation Details}
We evaluated the proposed framework with the M\&Ms Dataset~\cite{campello2021multi}, which provided the annotations of 3 cardiac structures, including the left ventricle (LV), right ventricle (RV) and the left ventricular myocardium (MYO).
The whole dataset gathered 320 Cardiac Magnetic Resonance (CMR) cases from 4 different scanner vendors, including 95 cases from Siemens (Vendor A), 125 cases from Philips (Vendor B), 50 cases from General Electric (Vendor C) and 50 cases from Cannon (Vendor D).
%

We resampled all volumes with unit spacing, cropped them centering at the heart region, and resized into $256\times 256$ pixels for pre-processing.
%
%
We conducted a three-fold cross validation, where the whole dataset was split into three proportions with the ratio of 4:4:3, and they took turns to be the testing data for evaluation.
Then, we randomly selected 50\% of the remaining data for training, and considered the rest of them as the validation set.
%
The implementation of the base generative model $G_B$ closely followed CGAN~\cite{isola2017image}, which is a commonly used paired image translation model.
The segmentation network backbone follows the recent work in multi-site learning~\cite{liu2020ms}, which is basically a UNet~\cite{ronneberger2015u} architecture attached with several residual blocks.
We adopted the Adam optimizer with the learning rate of 2e-4 to train this network.
We empirically set hyper-parameters $\lambda_{r}$, $\lambda_{d}$, $k$, $N$ as 0.5, 0.1, 3, 100, respectively.
We conducted the domain-sensitive feature whitening after each max-pooling layer in the backbone following~\cite{pan2018two}, and calculated the average of the feature whitening loss for each layer.
%
%
%
\begin{figure}[t]
	\centering
	\includegraphics[width=0.48\textwidth]{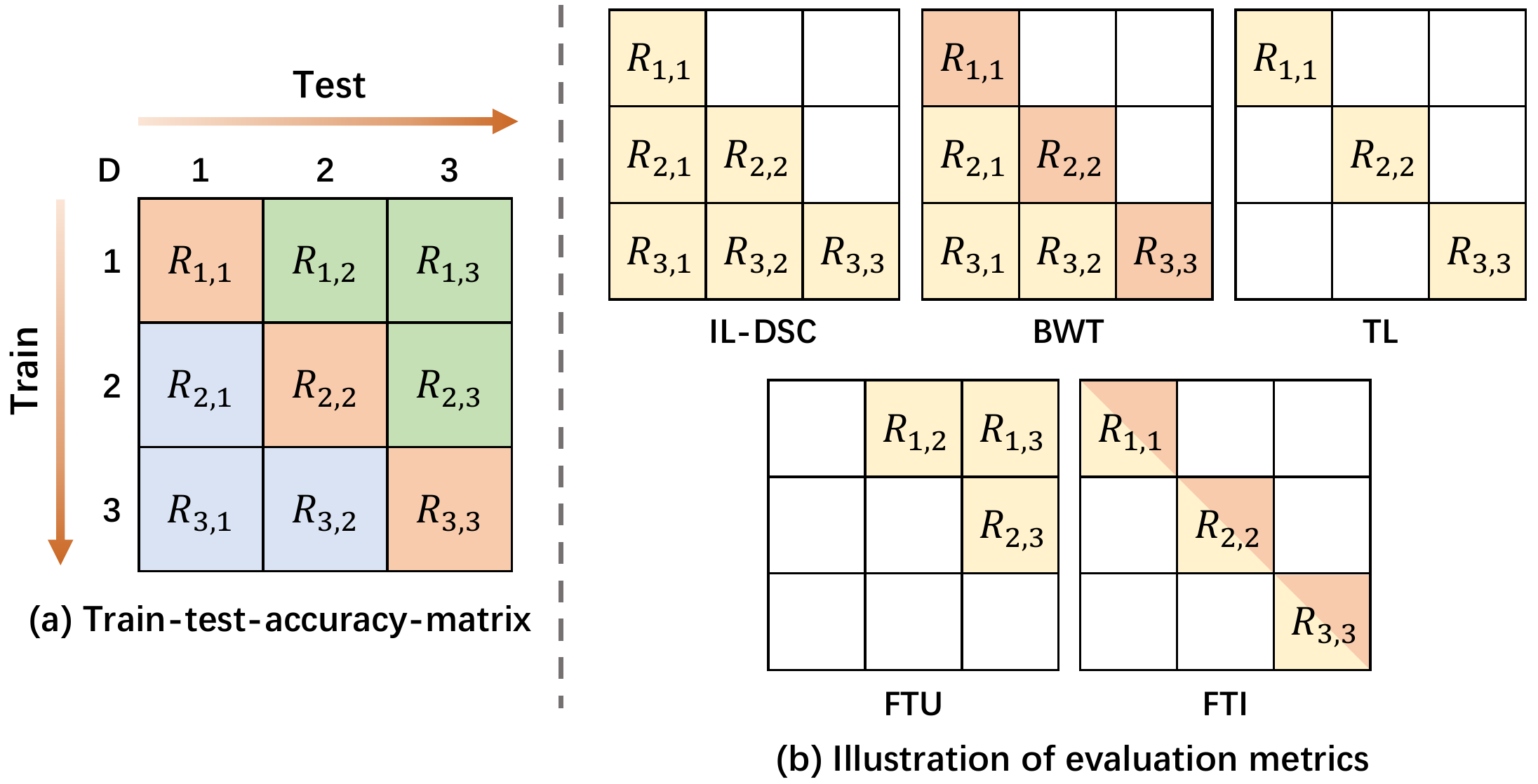}
	\caption{Evaluation metrics for the intermediate results in the entire incremental learning time span. (a): the illustration of train-test-accuracy-matrix. (b): For the example of taking DSC as the base metric, we use yellow to denote the elements of $R$ selected to calculate, and salmon orange to represent the contrastive items to compare with the items we picked.}
	\label{fig:eva-metrics}
\end{figure}

\subsection{Evaluation Metrics}
We employed Dice Similarity Coefficient (DSC) and 95\% Hausdorff Distance (HD) as the base metric.
To extensively evaluate incremental learning performance, we not only concentrated on the results in the last time step, but also considered all intermediate results throughout the incremental learning period for comprehensive assessments.
%
%
In the last time step, we computed the DSC scores and HD distances of past domains, current domains and unseen domains to measure its ability to alleviate catastrophic forgetting and generalization.
Furthermore, to evaluate the overall performance in the entire incremental learning time span, we employed the train-test-accuracy-matrix $R \in \mathcal{R}^{T\times T}$ following recent works~\cite{ozgun2020importance,yun2020tackling}.
When evaluated by DSC, each entry $R_{i, j}$ represents the DSC of the model tested on the $j$-th dataset after consecutively training from the $1$st dataset to the $i$-th dataset.
Similar rules also applied when taking HD for evaluation.
%
%
On top of it, we adopted 5 evaluation metrics to assess incremental learning methods from different aspects (See Fig.~\ref{fig:eva-metrics}).
%
Specifically, Backward Transfer (BWT) concentrates on the stability of the model, \ie, how well the model would retain previously acquired knowledge to prevent catastrophic forgetting.
If evaluated by DSC, BWT is computed as 
\begin{equation}
	BWT=\frac{2 \sum_{i=2}^{T} \sum_{j=1}^{i-1} 1-\left|\min \left(R_{i, j}-R_{j, j}, 0\right)\right|}{T(T-1)}.
\end{equation}
When evaluated with HD, we would formulate it in another format following Ozgun \etal~\cite{ozgun2020importance}, since HD normally do not have the maximum value.
We refer to this metric as $\text{BWT}^{+}$ for clarification and formulated as 
\begin{equation}
	\text{BWT}^{+}=\frac{2 \sum_{i=2}^{T} \sum_{j=1}^{i-1} \max \left(R_{i, j}-R_{j, j}, 0\right)}{T(T-1)},
\end{equation}
Transfer Learning (TL) focuses on the plasticity of the model, \ie, how well the model could adapt to the currently available dataset, which is calculated as the average of the diagonal elements in $R$.
Incremental Learning DSC (IL-DSC or IL-HD) measures the overall performance of backward transfer to past data and transfer learning on current data, which is calculated as the average of lower-triangular elements of $R$.
Forward Transfer Interference (FTI) also measures model plasticity but in a different way, which computes the performance differences between an incremental learning model and an individually trained model when testing on current dataset as follows:
\begin{equation}
	FTI=\frac{\Sigma_{i=2}^{T}(R_{i, i} - IT_{i, i})}{T-1},
\end{equation}
where $IT_{i, i}$ denotes the performance of the model individually trained on the $i$-th dataset and tested on the $i$-th dataset.
Forward Transfer on Unseen domains (FTU) evaluates the generalization ability of the model on unknown domains, which is calculated as the average of the upper-triangular elements in $R$.

\subsection{Incremental Learning with Single Domain}
%
%
\subsubsection{Experimental setting}
We first evaluated our method on a common scenario, where one vendor (domain) of data would be delivered at each time step and all four vendors of data in M\&Ms Dataset have consecutively arrived in alphabetical order over time.
As aforementioned, at each time point, only the currently arrived domain is accessible for training.
The datasets delivered in the past are no longer available for data privacy concerns.
Here, the total time step is set to 3 to enable the assessment on alleviating catastrophic forgetting on past domains, and the generalization to current and unseen domains at the same time.
Here, Vendor A and B are previously seen domains to observe the performance in retaining early acquired knowledge.
Meanwhile, Vendor C is the currently available domain to explore how well the model would accommodate to new information.
As Vendor D remains unknown and unused during the entire training process, we regarded it as unseen domain to further investigate the generalization potential to the data that might be encountered in future.

\subsubsection{Compared methods} 
For comparison methods, as a starter, we trained a separate segmentation model for each vendor (referred as Individual Training) to help observe the performance gains of forward knowledge transfer.
Then, we combined the data of all training vendors (\ie, Vendor A, B and C) and jointly trained a model.
Since it has full access of the training data arrived over time, while incremental learning methods could only access currently delivered one, it serves as the upper bound of incremental learning approaches.
%
Besides the straightforward incremental learning approach, \ie, sequentially finetune the model, we also compared with regularization-based methods including MAS-LR~\cite{ozgun2020importance} and EWC~\cite{baweja2018towards}, expansion-based methods like Orc-MML~\cite{gonzalez2020wrong}, and replay-based methods including LwF~\cite{li2017learning} and GR~\cite{shin2017continual}.

\subsubsection{Comparison results}
Table~\ref{table:single} reports the comparison results.
For the past domain performance, since Vendor A arrived earlier than Vendor B, the data of Vendor A would be easier to forget and exhibit more intense performance degradation.
%
With the proposed style-oriented replay module, our method could effectively mitigate the forgetting on past domains with the improvements of $1.01\%$ in the DSC of Vendor A, $0.95\%$ in BWT, and $1.02\%$ in IL-DSC, compared to the state-of-the-art approach.
For the performance of the current domain, as all incremental learning approaches could take full advantage of the annotations, most of them did not show significant performance differences between each other, although our framework outperformed others on the DSC of Vendor C and TL.
When it comes to forward transfer learning, \ie, how previously acquired knowledge would benefit the learning on current domain, our approach would lead to positive forward transfer with $1.13\%$ in FTI, showing the effectiveness to perform domain-sensitive feature whitening during optimization.
This operation also facilitates the generalization of unseen domains, where our framework achieved improved DSC performance of Vendor D and FTU.
%
%
%
%

\begin{figure*}[tp]
	\centering
	\includegraphics[width=0.85\textwidth]{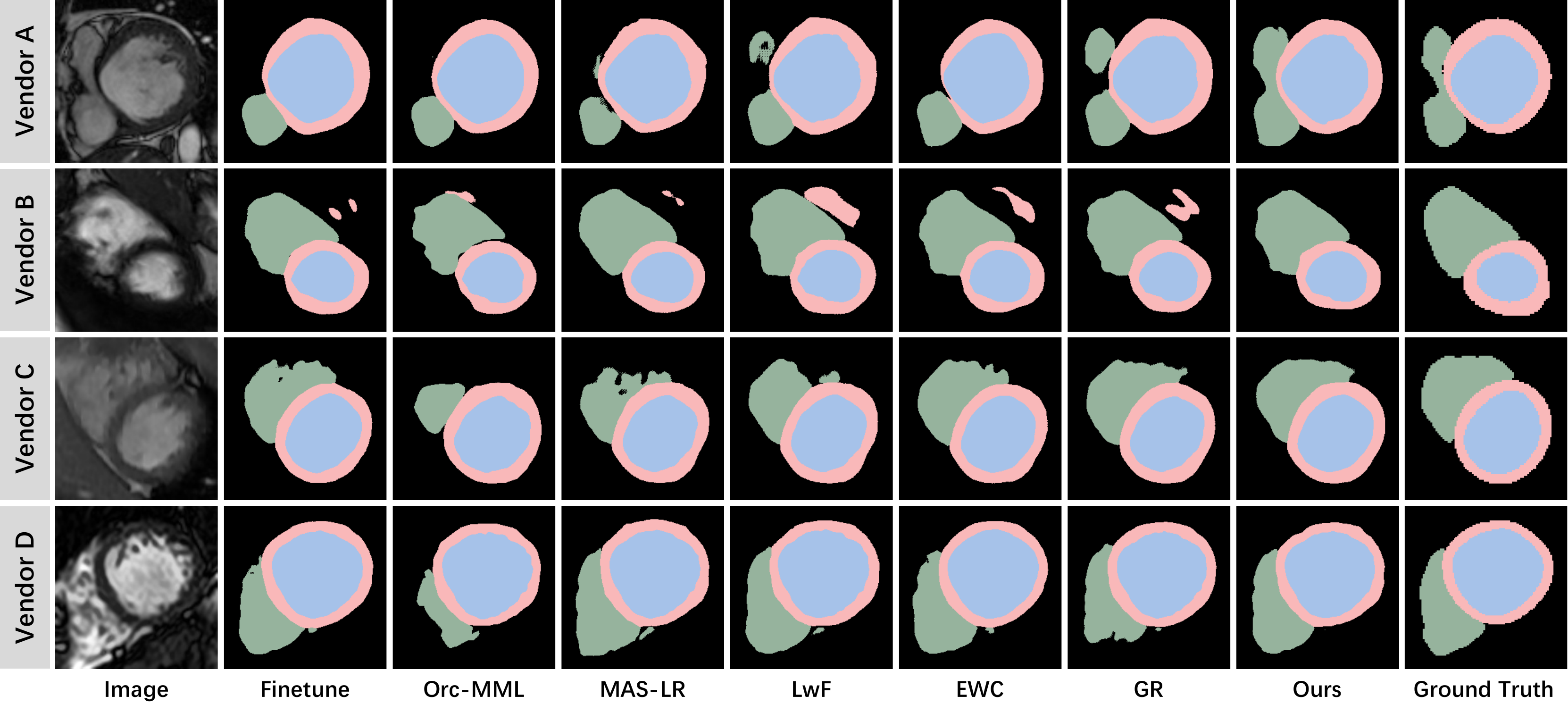}
	\caption{Visual comparisons for the methods incrementally learning from single domain. Here we use difference colors to denote different cardiac structures, including blue for LV, pink for MYO and green for RV. As observed, our predictions are more similar to the ground truth.}
	\label{fig:exp-results}
\end{figure*}

We also provided visual comparisons in Fig.~\ref{fig:exp-results}, where our segmentation results are less noisy and contain less false positive predictions.
Our framework could better partition the structures with ambiguous boundaries, \eg, the LV boundary in Vendor B, C and D.

\subsection{Incremental Learning with Compound Domain}
\subsubsection{Experimental setting}
We further explore a more complicated yet more practical scenario, where we loosen up on the assumption of data composition in each time step, \ie, these samples could all come from the same vendor (\ie, single domain) or could be compounded by two or more vendors (\ie, compound domain).
In practice, when the delivered datasets have insufficient samples (\eg, less than 5 volumes of data are collected from each site), it is common to merge these small datasets into one compound to build a suitable-size data collection (more than 10 volumes to prevent overfitting).
In our case, each dataset in the first two time steps (\eg, Vendor A or Vendor B) has around 100 samples.
However, each dataset delivered in the last two time steps (\eg, Vendor C or Vendor D) contains only 50 samples, which is almost half of previously arrived ones.
Naturally, one would merge the last two datasets as the current domain so that it would have a comparative size with past domains.
Here, we also set the total time step as 3 similar to the single-domain setting, and the vendors in M\&Ms Dataset have arrived in alphabetical order.
Moreover, we let Vendor C and Vendor D arrive at the same time step, and mix them up to form a compound dataset as the current domain.

\begin{table*}[th]
	\centering
	\caption{
		The Comparison Results to Incrementally Learning the Model by Both Compound Domain and Single Domain Data Over Time. We Reported the Performance at the Last Time Step and Also the Performance in the Entire Incremental Learning Lifespan in the Last Five Columns. For the results Evaluated by DSC, the Higher the Better. As For the Results Evaluated by HD, the Lower the Better. The Best Results Have Been Highlighted with \textbf{bold}}.
	\begin{tabular}{l|c|c|c|c|c|c|c|c} 
		\toprule[1pt]
		\multirow{2}{*}{Methods}                                               & \multicolumn{3}{c|}{DSC [\%]}   & \multirow{2}{*}{IL-DSC} & \multirow{2}{*}{BWT} & \multirow{2}{*}{TL} & \multirow{2}{*}{FTU} & \multirow{2}{*}{FTI}  \\ 
		\cline{2-4}
		& Vendor A & Vendor B & Vendor C\&D &                         &                      &                     &                      &                       \\ 
		\toprule[1pt]
		Individual Training                                                    & 88.91  & 89.82  & 87.46   & 87.28\%                 & 96.66\%              & 88.73\%             & 84.65\%              & N/A                   \\ 
		\hline
		\begin{tabular}[c]{@{}l@{}}Joint Training\\ (Upper bound)\end{tabular} & 89.21  & 90.56  & 87.83   & 89.64\%                 & 99.87\%              & 89.07\%             & 84.46\%              & 0.27\%                \\ 
		\midrule
		Sequentially Finetune                                                  & 85.04  & 89.51  & 88.22  & 88.08\%                 & 97.13\%              & 89.45\%             & 84.64\%              & 0.62\%                \\ 
		
		Orc-MML~\cite{gonzalez2020wrong}                                                                & 85.57  & 88.14  & 84.87   & 87.62\%                 & 97.84\%              & 87.84\%             & 84.76\%              & -1.69\%               \\ 
		
		MAS-LR~\cite{ozgun2020importance}                                                                 & 84.43  & 89.11  & 87.53   & 87.53\%                 & 87.69\%              & 88.46\%             & 84.35\%              & -0.31\%               \\ 
		
		LwF~\cite{li2017learning}                                                                    & 84.49  & 89.47  & 87.85   & 88.24\%                 & 97.54\%              & 89.23\%             & 84.86\%              & 0.27\%                \\ 
		
		EWC~\cite{baweja2018towards}                                                                    & 83.97  & 89.34  & 87.27   & 87.68\%                 & 96.86\%              & 88.92\%             & 84.69\%              & 0.23\%                \\ 
		
		GR~\cite{shin2017continual}                                                                     & 84.79  & 89.56  & 87.24  & 88.03\%                 & 97.67\%              & 88.79\%             & 84.73\%              & -0.36\%               \\ 
		\midrule
		Ours                                                                   & \textbf{88.04}  & \textbf{90.13}  & \textbf{88.76}   & \textbf{89.57\%}                & \textbf{99.34\%}              & \textbf{89.67\% }            & \textbf{85.22\%}              & \textbf{0.97\%}                \\
		\bottomrule[1pt]
	\end{tabular}
	\\[4pt]
	\begin{tabular}{l|c|c|c|c|c|c|c|c} 
		\toprule[1pt]
		\multirow{2}{*}{Methods}                                               & \multicolumn{3}{c|}{HD [Pixel]}                                                                & \multicolumn{1}{c|}{\multirow{2}{*}{IL-HD}} & \multicolumn{1}{c|}{\multirow{2}{*}{$\text{BWT}^{+}$}} & \multicolumn{1}{c|}{\multirow{2}{*}{TL}} & \multicolumn{1}{c|}{\multirow{2}{*}{FTU}} & \multicolumn{1}{c}{\multirow{2}{*}{FTI}}  \\ 
		\cline{2-4}
		& \multicolumn{1}{l|}{Vendor A} & \multicolumn{1}{l|}{Vendor B} & \multicolumn{1}{l|}{Vendor C\&D} & \multicolumn{1}{l|}{}                       & \multicolumn{1}{l|}{}                      & \multicolumn{1}{l|}{}                    & \multicolumn{1}{l|}{}                     & \multicolumn{1}{l}{}                      \\ 
		\toprule[1pt]
		Individual Training                                                    & 14.64                         & 9.12                          & 11.64                          & 10.99                                       & 3.54                                       & 9.65                                     & 13.57                                     & \multicolumn{1}{l}{N/A}                   \\ 
		\hline
		\begin{tabular}[c]{@{}l@{}}Joint Training\\ (Upper bound)\end{tabular} & 9.02                          & 7.60                          & 11.04                          & 8.99                                        & 0.00                                       & 9.47                                     & 13.53                                     & -0.26                                      \\ 
		\midrule
		Sequentially Finetune                                                  & 12.73                         & 8.77                          & 11.07                          & 10.41                                       & 2.51                                       & 9.48                                     & 13.62                                     & -0.25                                      \\ 
		
		Orc-MML [6]                                                               & 13.76                         & 8.43                          & 12.70                          & 10.54                                       & 2.21                                       & 10.18                                    & 13.71                                     & 0.81                                       \\ 
		
		MAS-LR [25]                                                                & 13.12                         & 9.27                          & 11.55                          & 10.57                                       & 1.86                                       & 10.12                                    & 13.86                                     & 0.71                                       \\ 
		
		LwF [9]                                                                    & 14.21                         & 9.95                          & 12.17                          & 11.09                                       & 3.53                                       & 9.83                                     & 13.65                                     & 0.28                                       \\ 
		
		EWC [26]                                                                    & 13.72                         & 9.28                          & 11.17                          & 10.67                                       & 2.86                                       & 9.59                                     & 13.64                                     & -0.09                                      \\ 
		
		GR [18]                                                                    & 12.57                         & 8.24                          & 11.53                          & 10.14                                       & 1.74                                       & 9.74                                     & 13.62                                     & 0.41                                       \\ 
		\midrule
		Ours                                                                   & \textbf{10.43}                         & \textbf{7.87}                          & \textbf{10.41}                          & \textbf{9.48}                                        & \textbf{0.82}                                       & \textbf{9.29}                                     & \textbf{12.37}                                     & \textbf{-0.65}                                      \\
		\bottomrule[1pt]
	\end{tabular}
	\label{table:compound}
\end{table*}

\subsubsection{Comparison results}
We reported the experiment results in Table~\ref{table:compound}.
Most existing works focused on a simple and rudimentary setting, \ie, the dataset arrived at each time step comes from one single site, to meet the basic needs of incremental learning.
Despite those comparison methods conducting additional operations to prevent forgetting, when it comes to a more practical yet more challenging setting, \ie,  compound-domain incremental learning, these operations are not always working and may even bring unwanted distraction to this process.

For example, no forgetting would happen for Orc-MML~\cite{gonzalez2020wrong}, if the domain classifier could correctly categorize every test sample.
Unfortunately, considering the compound domain is mixed with two distinguished data sources, it would make the VAE-based domain classifier extremely challenging to precisely capture their distinctive features, leading to less accurate judgements on domain identity.
Once the domain classifier produced inaccurate domain categories, the segmentation performance would experience more severe degradation in the compound-domain settings.
Let us say we have a test sample from Vendor A, but wrongly retrieved Vendor-B-trained or Vendor-C\&D-trained segmentation network for testing.
Then, the segmentation results would suffer from more forgetting due to triple domain discrepancies, \eg, the shifts from Vendor C to A, from Vendor D to A, and from Vendor B to A.
For the methods like LwF~\cite{li2017learning} and GR~\cite{shin2017continual}, at the time step $t$, they only reproduced and relearned the domain in last time step $D_{t-1}$ and missed out all domains before that, \eg, $D_{1}, ...., D_{t-2}$.
Sooner or later, the first arrived data (\eg, Vendor A in our case) will be utterly forgotten as time goes by.
Similar to Orc-MML~\cite{gonzalez2020wrong}, the segmentation performance on this vendor would also hugely degrade due to the triple domain shifts as mentioned before.
Regularization-based methods~\cite{ozgun2020importance,baweja2018towards} proposed to constrain large updates on the parameters that are important to previous domains.
However, as the compound domain samples come from two distinctive data sources, the constrained parameters selected by one vendor may not always completely overlap with the parameters determined by another vendor.
Certain non-overlapped parameters could be sacrificed for model convergence, leading to less satisfying performance in alleviating catastrophic forgetting.

%
As shown in Table~\ref{table:compound}, most existing methods suffered larger performance degradation on past domains, compared to learning from a single domain (see Table~\ref{table:single}).
Meanwhile, these methods produced worse DSC scores on current domain than the straightforward approach (\ie, Sequentially Finetune).
%
%
%
Armed with the proposed domain-sensitive feature whitening, our framework could better extend to the compound-domain setting, as we could effectively suppress the inference of domain-sensitive features to encourage it explore more on the domain-invariant ones.
Consequently, our approach achieved improved performance in the current domain and less forgetting on past domains, demonstrating the effectiveness of the proposed methods.

\subsection{Analysis of the key components}
\subsubsection{Effectiveness of Style-oriented Replay Module}
\begin{figure}[t]
	\centering
	\includegraphics[width=0.4\textwidth]{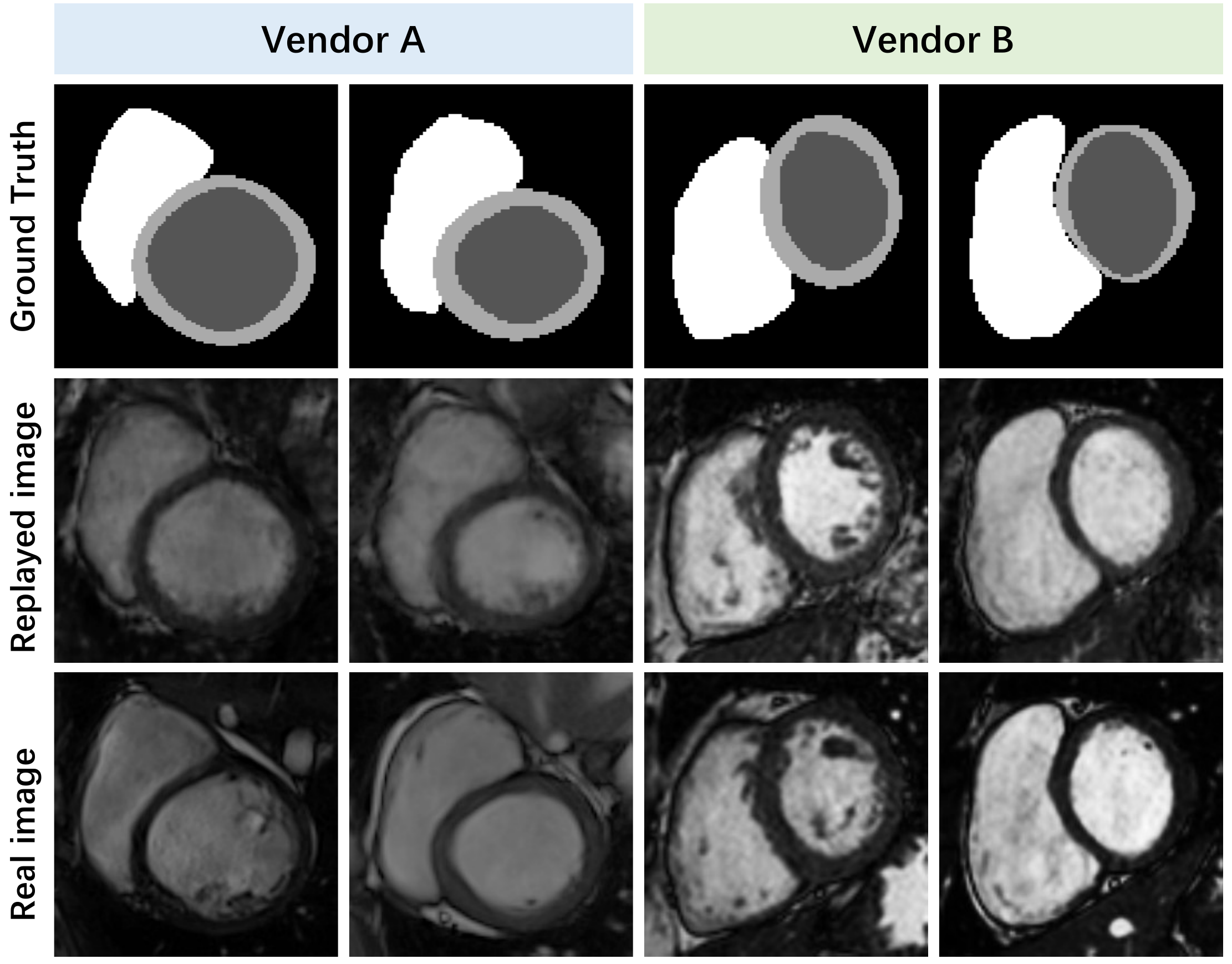}
	\caption{Visual comparisons between replayed images and real images.}
	\label{fig:replay}
\end{figure}

To evaluate the quality of replayed images, we first generate them by feeding the segmentation labels in the test set of Vendor A and B into the proposed style-oriented replay module, and directly compare the replayed images with the corresponding real images.
The visual comparisons are provided in Fig.~\ref{fig:replay}, where Vendor A is replayed by the base generative model $G_B$ and Vendor B is reproduced from a style-oriented generative model by modulating $G_B$ with style parameters $S_2$.
%
%
%
As shown in Fig.~\ref{fig:replay}, the structures of segmentation masks are well-preserved into the replayed images through the pixel-to-pixel mapping.
Meanwhile, the appearance in the replayed images are also similar to real ones.

Importantly, the proposed style-oriented replay module is memory-efficient.
Given the same architecture, storing one entire generative model for each dataset would cost around 156.6MB memory space.
In contrast, saving the style modulation parameters as we proposed only requires around 10MB space, greatly reducing memory consumption.

%
%

%

\subsubsection{Analysis of Domain-sensitive Feature Whitening}
%

As aforementioned, we employed $k$-means to cluster the variance matrix elements.
Since we selected the items in the $k$-th cluster as the domain-sensitive ones, different values of $k$ would lead to different ratios of suppressed items to all items.
To obtain the optimal $k$, we
observed the DSC changes of all training vendors given different values in single-domain incremental learning setting, as reported in Table~\ref{table:abl-k}.
When $k$ equals 3, our framework achieved the highest DSC scores on past domains, suggesting it is the optimal value.
In addition, we investigated the results of applying domain-sensitive feature whitening in different layers.
As pointed out by Pan~\etal~\cite{pan2018two}, shallow layers normally concentrate more on domain-specific style features than deep layers.
%
%
Inspired by that, we performed domain-sensitive feature whitening after each max-pooling layer in the segmentation backbone, and reported the results in Table~\ref{table:abl-layer}.
As observed, conducting feature whitening for any max-pooling layer would lead to improved performance, and appending it after the first one is most beneficial.
When concurrently performing domain-sensitive feature whitening after all max-pooling layers, more improvements could be achieved.

\begin{table}[th]
	\centering
	\caption{
		Analysis of the Number of Clusters in Domain-sensitive Feature Whitening. Here, We Change the Values of $k$ to Adjust the Suppression Ratio, from None Suppression to Complete Suppression, and Report the DSC Scores of All Vendors to Select the Optimal $k$ Value.
	}
	\begin{tabular}{c|c|c|c|c|c} 
		\toprule[1pt]
		$k$   & Ratio    &Vendor A &Vendor B &Vendor C &Vendor D  \\ 
		\midrule
		N/A & 100.00\% & 87.49\%  & \textbf{90.15\%}  & 87.44\%  & 85.89\%   \\
		
		2   & 90.61\%   & 87.34\%  & 89.58\%  & 87.78\%  & 85.59\%   \\ 
		
		3   & 76.40\%  & \textbf{87.95\%}  & 89.73\%  & 87.51\%  & \textbf{86.17\%}   \\ 
		
		4   & 65.38\%  & 87.47\%  & 89.66\%  & 87.82\%  & 86.04\%   \\ 
		
		5   & 52.23\%  & 87.55\%  & 89.75\%  & 87.46\%  & 85.97\%   \\ 
		
		10  & 24.15\%  & 86.78\%  & 89.41\%  & \textbf{87.97\%}  & 85.86\%   \\ 
		
		15  & 12.30\%  & 86.87\%  & 89.58\%  & 87.82\%  & 85.51\%   \\ 
		
		20  & 8.46\%  & 86.86\%  & 89.33\%  & 87.69\%  & 85.77\%   \\ 
		
		N/A & 0.00\%   & 87.44\%  & 89.57\%  & 86.82\%  & 85.36\%   \\ 
		\bottomrule
	\end{tabular}
	\label{table:abl-k}
\end{table}

\begin{table}[th]
	\centering
	\caption{
		Analysis of Applying Domain-sensitive Feature Whitening in Different Layers.
	}
	\begin{tabular}{c|c|c|c|c} 
		\toprule[1pt]
		Layer  & Vendor A         & Vendor B         & Vendor C          & Vendor D          \\ 
		\midrule
		None   & 87.44\%          & 89.57\%          & 86.82\%           & 85.36\%           \\ 
		\midrule
		Pool-1 & 87.67\%          & 89.51\%          & 87.33\%           & 85.89\%           \\ 
		
		Pool-2 & 87.52\%          & 89.42\%          & 87.17\%           & 85.63\%           \\ 
		
		Pool-3 & 87.46\%          & 89.59\%          & 87.26\%           & 85.61\%           \\ 
		
		Pool-4 & 87.49\%          & 89.53\%          & 87.15\%           & 85.74\%           \\ 
		\midrule
		All    & \textbf{87.95\%} & \textbf{89.73\%} & \textbf{87.51\% } & \textbf{86.17\%}  \\
		\bottomrule[1pt]
	\end{tabular}
	\label{table:abl-layer}
\end{table}

\section{Discussion}

In the past several years, deep learning models have demonstrated satisfying performance in various applications.
Most of them were obtained with static training, \ie, gathering an annotated dataset with suitable size and optimizing a network once for all.
However, as time goes by, new data keeps pouring in and new requirements may arise.
%
Similar to the Apps in our cellphones or the operating systems in our computers, which constantly update themselves with fixed bugs and improved functions, the models built for medical applications also have essential needs to iteratively upgrade and gradually evolve~\cite{li2017learning,saha2020gradient}.
For example, a company once collected a batch of data from one hospital and trained a deployable model on it.
As time goes by, more collaborations have been established where the second or more hospitals decide to participate and offer their private data for model training.
Since multi-site data often contain domain discrepancy among each other, inevitably, the previously trained model needs to incrementally update itself to remain viable to deploy in the first hospital while well adapting to current hospitals.
In this regard, the desirable deep model should incrementally learn and adapt over time.
Our work proceeds along this promising direction and presents a domain-incremental learning framework for cardiac image segmentation.
Equipped with our style-oriented replay module and domain-sensitive feature whitening, our framework could effectively alleviate catastrophic forgetting on previously seen datasets, without harming the potential to generalize to other datasets that might be encountered in future.

When it comes to how to prevent catastrophic forgetting on previously learnt data, most existing works made a tradeoff by allowing larger storage usages~\cite{gonzalez2020wrong, saha2020gradient} or sub-optimal generalization to new datasets or unseen datasets~\cite{ozgun2020importance,baweja2018towards,kirkpatrick2017overcoming}.
%
%
%
In the medical field, the former (\ie, larger memory consumption) is more acceptable, since medical applications are usually sensitive and demanding to model accuracy, where even tiny mistakes may have serious consequences~\cite{tarroni2020large}.
In the meantime, considerable efforts have been devoted to minimizing the required memory augmentation for each new task or dataset with memory-efficient incremental learning approaches~\cite{saha2020gradient,yoon2019scalable}.
Our framework also intends to optimize the memory storage of each new domain, where we maximally reuse the parameters of the base generative model and only save the specific style modulation parameters ($\approx10$MB) instead of the whole generative model ($\approx156$MB).
%
%
With recent advances in model compression and pruning~\cite{wu2021spirit},
we believe the memory consumption could be further reduced in future.

%
Besides that, our framework concentrated on one major cause of forgetting, \ie, the domain shifts caused by appearance, and assumed no output variability across vendors.
However, different diseases often have distinctive abnormal heart structures.
Leaving the variability of heart shapes out of consideration could still bring a certain degree of forgetting.
Since the M\&Ms dataset does not provide the sample-wise disease information, \ie, the disease type of each sample, it is hard to quantitatively evaluate its impact. We hope to solve it in the future.
It is worth to mention that since we involve all replayed past datasets to help the model remember previously learned knowledge via $\mathcal{L}_{sup}^{t}$, it would bring certain cost on training time as time goes on.
The time cost of $\mathcal{L}_{sup}^{t}$ not only relates to how many datasets need to be replayed (\ie, $t-1$ past datasets at the current time step $t$), but also depends on the size of current incoming dataset (\ie, the scale of $D_t$).
In our experiment, for each epoch, the average time cost increases of each incremental learning step are around 1 min in the single-domain and compound-domain settings, using one NVIDIA TITAN XP GPU card.

\section{Conclusion}

We presented a domain-incremental learning framework to enable progressive updates on cardiac image segmentation model by investigating sequentially-arrived heterogenous datasets.
We first presented a style-oriented replay module to recover the past domain inputs in a memory-efficient and structure-realistic way, and regularly reminded the model with replayed past domains to mitigate catastrophic forgetting.
Furthermore, we proposed to perform domain-sensitive feature whitening during optimization to suppress the dependency on the features that sensitively respond to domain changes and encourage it to leverage domain-invariant content ones, to facilitate the generalization on current domains and unseen domains.
We conducted extensive experiments on M\&Ms Dataset in single-domain and compound-domain incremental learning settings to validate the superiority of our method.
\bibliographystyle{IEEEtran}
\bibliography{0-ref}

\end{document}